\pdfoutput=1

\documentclass[11pt]{article}

\usepackage[]{acl}

\usepackage{times}
\usepackage{latexsym}
\usepackage{enumitem}
\usepackage{multirow}
\usepackage[T1]{fontenc}

\usepackage[utf8]{inputenc}

\usepackage{microtype}

\usepackage{inconsolata}

\usepackage{graphicx}

\usepackage{tcolorbox}
\usepackage{lipsum}
\usepackage{hyperref}
\usepackage{url}
\usepackage{booktabs} 
\usepackage{arydshln}
\usepackage{multicol}
\usepackage{ulem}
\usepackage{subcaption}
\usepackage{amsmath}
\usepackage{amssymb}
\usepackage{pifont}
\usepackage{amsmath}
\usepackage{stfloats}

\usepackage{tikz}
\newcommand{\bgf}[1]{\tikz[baseline=(X.base)]{\node(X)[rectangle, fill=blue!10, rounded corners, text height=1.4ex,text depth=-1.0ex,draw=white]{#1};}}
\newcommand{\bgfr}[1]{\tikz[baseline=(X.base)]{\node(X)[rectangle, fill=purple!10, rounded corners, text height=1.4ex,text depth=-1.0ex]{#1};}}

\DeclareRobustCommand{\mbzuai}{%
  \begingroup
  \vspace{0em}%
  \raisebox{0em}{%
  \includegraphics[height=1em]{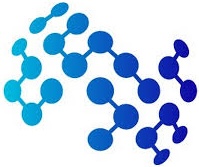}%
  }%
  \kern 0em%
  \endgroup
}

\title{Balanced Multi-Factor In-Context Learning for Multilingual Large Language Models}

\author{Masahiro Kaneko \quad
        Alham Fikri Aji \quad
        Timothy Baldwin \\
        \mbzuai{} MBZUAI \\
        {\tt \{Masahiro.Kaneko, Alham.Fikri, Timothy.Baldwin\}@mbzuai.ac.ae}}

\begin{document}
\maketitle
\begin{abstract}

Multilingual large language models (MLLMs) are able to leverage in-context learning (ICL) to achieve high performance by leveraging cross-lingual knowledge transfer without parameter updates.
However, their effectiveness is highly sensitive to example selection, particularly in multilingual settings.
Based on the findings of existing work, three key factors influence multilingual ICL: (1) semantic similarity, (2) linguistic alignment, and (3) language-specific performance. 
However, existing approaches address these factors independently, without explicitly disentangling their combined impact, leaving optimal example selection underexplored.
To address this gap, we propose balanced multi-factor ICL (\textbf{BMF-ICL}), a method that quantifies and optimally balances these factors for improved example selection.
Experiments on mCSQA and TYDI across four MLLMs demonstrate that BMF-ICL outperforms existing methods.
Further analysis highlights the importance of incorporating all three factors and the importance of selecting examples from multiple languages.

\end{abstract}

\section{Introduction}
\label{sec:itr}

Multilingual large language models (MLLMs) leverage cross-lingual knowledge transfer by learning from text in diverse languages~\cite{conneau2019cross,conneau-etal-2020-unsupervised,xue-etal-2021-mt5,Scao2022BLOOMA1}.
In-context learning (ICL) further enhances performance by selecting a small number of examples from candidate sets, leading to high accuracy on various tasks without parameter updates~\cite{liu-etal-2022-makes}.
A key approach to maximize the cross-lingual transfer for MLLMs is to select examples from multilingual candidate pools.
Since ICL performance heavily depends on which examples are chosen, the example selection strategy is crucial~\cite{NEURIPS2021_5c049256,zhao2021calibrate,lu-etal-2022-fantastically,koike-etal-2024-prompt,hida2024social,oba2024contextual}.

According to existing work on example selection in multilingual ICL, three main factors influence effectiveness: (1) \textbf{semantic similarity}, (2) \textbf{linguistic alignment}, and (3) \textbf{language-specific performance}.
Selecting examples semantically similar to the input often improves performance~\cite{nie-etal-2023-cross,tanwar-etal-2023-multilingual,liu-etal-2022-makes}.
Using examples from languages that are morphologically and grammatically similar to the target language can lead to stronger knowledge transfer~\cite{johnson-etal-2017-googles,pires-etal-2019-multilingual,yamashita2020cross,winata-etal-2022-cross,dolicki2021analysing}. 
Furthermore, inference performance varies by language, and leveraging data from high-resource languages like English can boost results in low-resource languages~\cite{winata-etal-2021-language,etxaniz-etal-2024-multilingual}.

\begin{figure}[!t]
  \centering
  \includegraphics[width=0.5\textwidth]{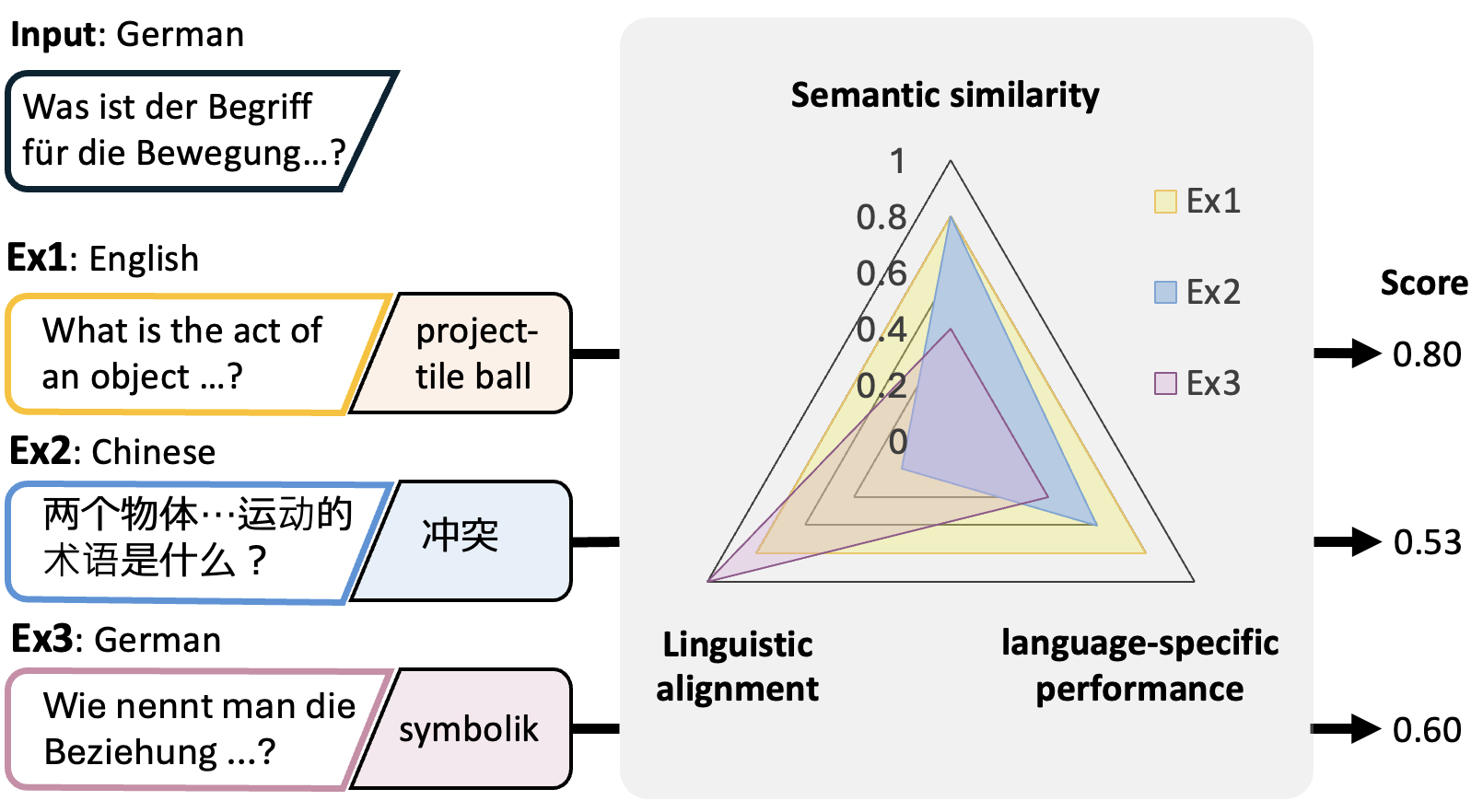}
  \caption{Our proposed method, BMF-ICL, selects multilingual examples for ICL by considering three factors: semantic similarity, linguistic alignment, and language-specific performance.}
  \label{fig:abst}
\end{figure}

While these three factors are integral to multilingual ICL, existing research typically does not combine them together. 
Moreover, existing work has two limitations that prevent them from combining these factors: the lack of quantified selection criteria, and the absence of explicit differentiation among the factors.
The languages, for example, are often selected heuristically, either based on language groups and geographic regions for linguistic alignment~\cite{nguyen-etal-2024-democratizing,winata-etal-2022-cross}, or based on per-language data sizes in MLLM training data for language-specific performance~\cite{winata-etal-2021-language,nie-etal-2023-cross}.
Additionally, existing studies typically use multilingual sentence embeddings~\cite{conneau-etal-2020-unsupervised} that do not explicitly distinguish between semantic similarity and linguistic alignment, making it impossible to optimize the balance between them~\cite{nie-etal-2023-cross}.

In this study, we propose a method called balanced multi-factor ICL (\textbf{BMF-ICL}), a method that defines explicit metrics for semantic similarity, linguistic alignment, and language-specific performance in multilingual in-context learning (MICL), then selects examples by optimally balancing these factors.
\autoref{fig:abst} presents an overview of BMF-ICL, which considers three scores for multilingual example selection.
Specifically, we quantify each factor as follows:
\begin{description}
    \item[(1) \textbf{Semantic similarity:}] We employ LaBSE ~\cite{feng-etal-2022-language}, a language-agnostic sentence embedding model to score the similarity between the input and candidate examples.
    \item[(2) \textbf{Linguistic alignment:}] We use lang2vec~\cite{littell-etal-2017-uriel}, which captures morphological and grammatical features, to assess how closely the input language aligns with the candidate language.
    \item[(3) \textbf{Language-specific performance:}] We compute the likelihood of producing the correct answer for each language when the MLLM is provided with the candidate example’s inputs.
\end{description}
To select examples while balancing these three scores, we take their weighted sum and optimize the weights on development data. 

We evaluate both existing approaches and our proposed method on two benchmark datasets, mCSQA~\cite{sakai-etal-2024-mcsqa} and TYDI~\cite{Clark2020TyDiQA}. 
The experimental results across four MLLMs demonstrate that BMF-ICL consistently achieves the highest accuracy compared to existing methods.  
Further analysis highlights the importance of considering all three factors jointly.  
Notably, in over 95\% of the cases, the proposed method selects examples from two or more languages, demonstrating the performance benefits derived from multilingual data.

\section{Balanced Multi-Factor In-Context Learning (BMF-ICL)}
\label{sec:micl}

\begin{figure}[!t]
  \centering
  \includegraphics[width=0.5\textwidth]{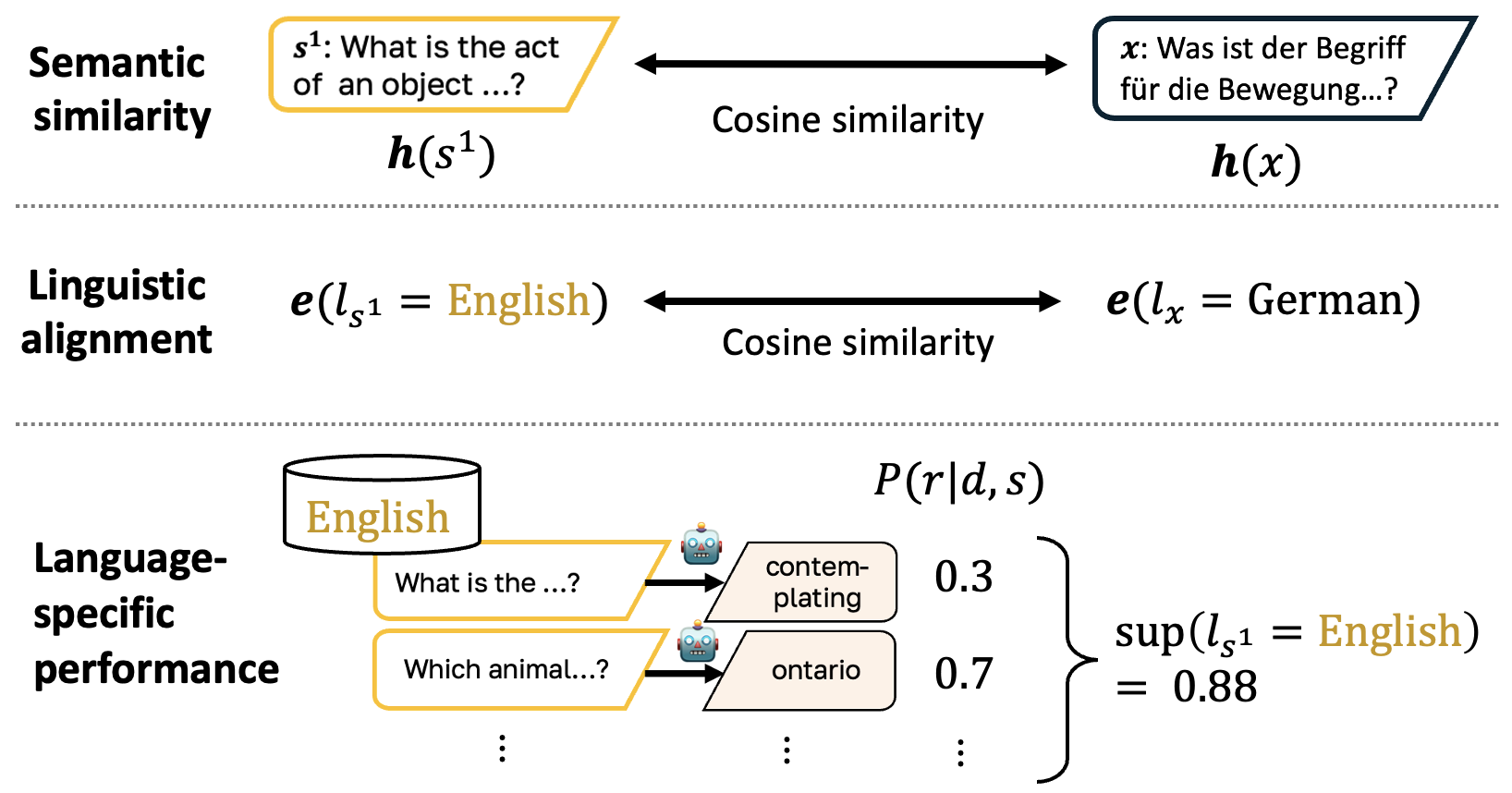}
  \caption{An overview of how BMF-ICL computes semantic similarity, linguistic alignment, and language-specific performance scores to select multilingual examples.}
  \label{fig:method}
\end{figure}

We first explain ICL, followed by a discussion of the proposed method for example selection, which takes into account the scores of semantic similarity, linguistic alignment, and language-specific performance, and comprehensively considers these factors.  
\autoref{fig:method} illustrates how BMF-ICL calculates these three scores.

\subsection{In-Context Learning}

Let $x$ be an input text and $y$ be an output text generated by the LLM with parameters $\theta$.
In ICL, given a task definition text $d$, the set of examples $\mathcal{E}$, and the input text $x$, the LLM generates $y$ by maximizing the following conditional probability:
\begin{equation}
    \label{eq:icl}
    y = \underset{\hat{y}}{\operatorname{argmax}} P(\hat{y} \mid d, \mathcal{E}, x;\theta)
\end{equation}

Our goal is to construct the set of examples $\mathcal{E}$ in a way that maximizes the model’s performance or the quality of the generated text. Specifically, $\mathcal{E}$ is composed of $k$ example pairs, drawn from a pool that contains source and reference texts ($\mathcal{S}$, $\mathcal{R}$).
\begin{align}
\mathcal{E} = \{(s^{(i)}, r^{(i)}) \in (\mathcal{S}, \mathcal{R})\}_{i=1}^{k}
\end{align}
Here, $s^{(j)}$ and $r^{(j)}$ ($j = 1, \dots, k$) are the top $k$ instances ranked by a selection method.

\subsection{Example Selection via Multi-Factor}
We propose a balanced approach to example selection by integrating three key factors for MICL.
Specifically, we select the top $k$ instances from the example pool to form $\mathcal{E}$ according to the highest weighted sum of the following three scores:
\begin{align}
\label{eq:total}
\operatorname{score^{(j)}} = \alpha \operatorname{score^{(j)}_\text{sem}} + \beta \operatorname{score^{(j)}_\text{lag}} + \gamma \operatorname{score^{(j)}_\text{per}},
\end{align}
where $\operatorname{score^{(j)}_\text{sem}}$ represents the semantic similarity between the input text and the source text, $\operatorname{score^{(j)}_\text{lag}}$ represents the linguistic similarity between the input text and the source text, and $\operatorname{score^{(j)}_\text{per}}$ reflects the model’s performance in generating the reference text from the source text in the target language.
The scalar coefficients $0 \leq \alpha, \beta, \gamma$ satisfy $\alpha + \beta + \gamma = 1$.

\paragraph{Semantic similarity}
To calculate the semantic similarity between input text $x$ and source texts $s^{(i)}$ in the example pool, let $\boldsymbol{h}(x)$ and $\boldsymbol{h}(s^{(j)})$ be their sentence embeddings.
We use LaBSE~\cite{feng-etal-2022-language}\footnote{\url{https://huggingface.co/sentence-transformers/LaBSE}} as the multilingual sentence embedding model.
LaBSE learns multilingual embeddings through contrastive learning over large-scale parallel data, enabling consistent semantic similarity computation across languages.
The semantic similarity score $\text{score}^{(j)}_{\text{sem}}$ for the $j$-th source text is then defined by the cosine similarity:
\begin{align}
\text{score}^{(j)}_{\text{sem}} = \text{cos}(\boldsymbol{h}(x), \boldsymbol{h}(s^{(j)})).
\end{align}

\paragraph{Linguistic alignment} 
To calculate the linguistic similarity, let $\boldsymbol{e}(l_x)$ and $\boldsymbol{e}(l_{s^{(j)}})$ be linguistic embeddings corresponding to the languages $l_x$ (the language of $x$) and $l_{s^{(j)}}$ (the language of $s^{(j)}$).
We use fasttext-langdetect~\cite{joulin2016bag,joulin2016fasttext}\footnote{\url{https://pypi.org/project/fasttext-langdetect/}} to detect the languages of the input texts and the source texts.
The linguistic embeddings $\boldsymbol{e}(l_x)$ and $\boldsymbol{e}(l_{s^{(j)}})$ are obtained from lang2vec~\citep{littell-etal-2017-uriel},\footnote{\url{https://www.cs.cmu.edu/~dmortens/projects/7_project/}} which encodes typological, geographical, and phylogenetic properties of languages.

The linguistic similarity score $\text{score}^{(j)}_{\text{lag}}$ for the $j$-th source text is defined as:
\begin{align}
\text{score}^{(j)}_{\text{lag}} = \text{cos}(\boldsymbol{e}(l_{x}), \boldsymbol{e}(l_{s^{(j)}})).
\end{align}

\paragraph{Language-specific performance}

Finally, we measure the model's performance in each language by evaluating how well it generates the reference text $r^{(j)}$ from the source text $s^{(j)}$.
For a target language $l_{\textrm{tgt}}$, we define the sub-dataset $(\mathcal{S}_{l_{\textrm{tgt}}}, \mathcal{R}_{l_{\textrm{tgt}}})$ as follows:
\begin{align}
(\mathcal{S}_{l_{\textrm{tgt}}}, \mathcal{R}_{l_{\textrm{tgt}}}) = \{(s', r' \in (\mathcal{S}, \mathcal{R}) \mid l_{s'} = l_{\textrm{tgt}} \}.
\end{align}

Here, $l_{\textrm{tgt}}$ can be any language present in the candidate examples.
We define the model’s inference ability for each language $\text{per}(l_{t})$ for language $l_{t}$ as the average log-likelihood of each reference text $r'$ given its corresponding source text $s'$:

{\small
\begin{align}
 \text{per}(l_{t}) = \frac{1}{|\mathcal{S}_{l_{t}}|} \sum_{(s',r') \in \mathcal{S}_{l_{t}}, \mathcal{R}_{l_{t}}} \frac{1}{|r'_{i}|} \sum_{i=1}^{|r'_{i}|} \log P(r'_{i} \mid d, s';\theta)
\end{align}
}

For the \(j\)-th source text in the example pool, the performance score \(\text{score}^{(j)}_{\text{per}}\) is given by:
\begin{align}
\text{score}^{(j)}_{\text{per}} = \text{per}(l_{s^{(j)}})
\end{align}

By combining these three scores in Eq.~\eqref{eq:total}, our method aims to select examples that simultaneously capture semantic similarity, linguistic alignment, and model performance, thereby improving the overall effectiveness of in-context learning.

\section{Experiments}
\label{sec:exp}

\subsection{Settings}
\paragraph{Dataset} 
Many multilingual datasets are constructed by translating a single-language dataset into multiple other languages, resulting in parallel content across languages. 
This setup diverges from realistic scenarios in which data distributions vary by language and also prevents the assessment of potential synergies gained from multilingual ICL.
Therefore, we use two multilingual datasets, each originally developed in its own language rather than through translation.

\textbf{mCSQA}~\citep{sakai-etal-2024-mcsqa}\footnote{\url{https://huggingface.co/datasets/yusuke1997/mCSQA}} contains multilingual commonsense question-answering data in a multiple choice format for 8 languages.
\textbf{TYDI}~\citep{Clark2020TyDiQA} is a question-answering dataset covering 11 typologically diverse languages.
We frame the task as gold passage generation, where both the context and question are provided, and the model is required to generate the answer.

\autoref{apx:tbl:data} in \autoref{apx:sec:dataset} shows the data size and language group for each language in mCSQA and TYDI.
In both datasets, we generate answers and evaluate them based on exact-match accuracy.

\paragraph{Model} 

We explore both open-weight and closed-weight models. Specifically, we use \textbf{BLOOMZ}~\citep{Scao2022BLOOMA1},\footnote{\url{https://huggingface.co/bigscience/bloomz}} \textbf{Aya}~\citep{ustun2024aya},\footnote{\url{https://huggingface.co/CohereForAI/aya-23-8B}} \texttt{gpt-3.5-turbo-0125} (\textbf{GPT-3.5})~\citep{Brown2020LanguageMA}, and \texttt{gpt-4-turbo-2024-04-09} (\textbf{GPT-4}) as multilingual LLMs.
We use eight NVIDIA A100 GPUs for our experiments.

\paragraph{ICL Setup}
We use the training sets from mCSQA and TYDI as example pools.
To determine the optimal prompt configuration, we vary the number of examples (2, 4, 8, and 16) and test four different prompts.
The prompts are based on existing research~\citep{Robinson2022LeveragingLL} and prompt guidelines.\footnote{\url{https://huggingface.co/docs/transformers/v4.37.0/en/tasks/prompting}}
Across these experiments, mCSQA and TYDI achieved the best performance with 8 examples using the following prompts:

\begin{tcolorbox}[fontupper=\ttfamily, title={Prompt for mCSQA}]
  \scriptsize
  Answer the question.

  Question: [Question of Example 1]

  a. [Choice A of Example 1]
  
  b. [Choice B of Example 1]
  
  c. [Choice C of Example 1]
  
  d. [Choice D of Example 1]
  
  e. [Choice E of Example 1]

  Answer: [Answer of Example 1]
  
  \vdots

  Question: [Question of Example 8]

  a. [Choice A of Example 8]
  
  b. [Choice B of Example 8]
  
  c. [Choice C of Example 8]
  
  d. [Choice D of Example 8]
  
  e. [Choice E of Example 8]

  Answer: [Answer of Example 8]

  Question: [Question of Input]

  a. [Choice A of Input]
  
  b. [Choice B of Input]
  
  c. [Choice C of Input]
  
  d. [Choice D of Input]
  
  e. [Choice E of Input]

  Answer: 
\end{tcolorbox}

\begin{tcolorbox}[fontupper=\ttfamily, title={Prompt for TYDI}]
  \scriptsize
  Answer the question using the context.
  
  Context: [Context of Example 1]

  Question: [Question of Example 1]

  Answer: [Answer of Example 1]

  Context: [Context of Example 2]

  Question: [Question of Example 2]

  Answer: [Answer of Example 2]
  
  \vdots

  Context: [Context of Example 7]

  Question: [Question of Example 7]

  Answer: [Answer of Example 7]

  Context: [Context of Example 8]

  Question: [Question of Example 8]

  Answer: [Answer of Example 8]

  Context: [Context of Input]

  Question: [Question of Input]

  Answer: 
\end{tcolorbox}

\paragraph{Weight Selection for BMF-ICL}
We explore all combinations of $\alpha$, $\beta$, and $\gamma$ in Eq.~\autoref{eq:total} from 0 to 1 in increments of 0.1, ensuring $\alpha + \beta + \gamma = 1$.
We then select the combination of weights that yield the best performance on the development sets of mCSQA and TYDI datasets.\footnote{We show specific weights in \autoref{apx:sec:weights}}
These selected weights, together with the 8-example prompt configuration, define our final method for BMF-ICL.

\begin{table*}[t!]
  \scriptsize
  \begin{subtable}{\textwidth}
  \centering
  \resizebox{0.9\textwidth}{!}{
  \begin{tabular}{llccccccccc}
    \toprule
     & & en & zh & fr & de & ja & nl & pt & ru \\
    \midrule
    \multirow{4}{*}{w/ TL} & Random-ICL & \textbf{60.3} & 38.2 & 58.5 & 64.0 & 44.7 & 59.7 & 65.5 & 32.2 \\
    & \citet{etxaniz-etal-2024-multilingual} & - & \bgfr{30.3} & \bgf{58.9} & \bgf{66.3} & \bgfr{38.7} & \bgf{60.4} & \bgfr{65.1} & \bgfr{31.5} \\
    & \citet{nguyen-etal-2024-democratizing} & - & \bgfr{35.5} & \bgf{58.7} & \bgf{65.9} & \bgfr{39.1} & \bgf{60.0} & \bgfr{64.0} & \bgfr{31.5} \\
    & BMF-ICL & \bgfr{60.0} & \bgf{\textbf{40.6}}$^{\dagger}$ & \bgf{\textbf{60.2}}$^{\dagger}$ & \bgf{\textbf{67.0}} & \bgf{\textbf{46.1}}$^{\dagger}$ & \bgf{\textbf{61.5}}$^{\dagger}$ & \bgf{\textbf{67.8}}$^{\dagger}$ & \bgf{\textbf{32.5}} \\
    \hdashline
    \multirow{5}{*}{w/o TL} & Non-ICL & 44.0 & 24.5 & 43.8 & 29.6 & 25.4 & 38.3 & 37.0 & 21.5 \\
    & \citet{winata-etal-2021-language} &  - & \bgf{26.1} & \bgf{48.8} & \bgf{\textbf{60.0}} & \bgf{30.1} & \bgf{50.2} & \bgf{43.9} & \bgf{23.6} \\
    & \citet{winata-etal-2022-cross} & \bgf{47.9} & \bgf{26.7} & \bgf{48.3} & \bgf{55.3} & \bgf{31.7} & \bgf{47.5} & \bgf{44.7} & \bgf{24.0} \\
    & \citet{nie-etal-2023-cross} & \bgf{49.6} & \bgf{28.8} & \bgf{50.1} & \bgf{58.2} & \bgf{35.5} & \bgf{46.9} & \bgf{47.3} & \bgf{24.4} \\
    & BMF-ICL & \bgf{\textbf{53.1}}$^{\dagger}$ & \bgf{\textbf{31.4}}$^{\dagger}$ & \bgf{\textbf{52.5}}$^{\dagger}$ & \bgf{59.3} & \bgf{\textbf{38.3}}$^{\dagger}$ & \bgf{\textbf{51.0}} & \bgf{\textbf{49.9}}$^{\dagger}$ & \bgf{\textbf{25.7}}$^{\dagger}$ \\
    \bottomrule
  \end{tabular}
  }
  \caption{BLOOMZ.}
  \end{subtable}
  \hfill
  \begin{subtable}{\textwidth}
  \centering
  \resizebox{0.9\textwidth}{!}{
  \begin{tabular}{llccccccccc}
    \toprule
    & & en & zh & fr & de & ja & nl & pt & ru \\
    \midrule
    \multirow{4}{*}{w/ TL} & Random-ICL & 61.1 & 37.1 & 59.0 & 63.9 & 46.6 & 60.1 & 65.0 & 33.6 \\
    & \citet{etxaniz-etal-2024-multilingual} & - & \bgfr{37.0} & \bgf{61.5} & \bgf{66.5} & \bgfr{46.0} & \bgf{\textbf{64.0}} & \bgf{66.7} & \bgfr{32.9}  \\
    & \citet{nguyen-etal-2024-democratizing} & - & \bgf{38.8} & \bgf{60.1} & \bgfr{62.9} & \bgfr{47.0} & \bgf{63.9} & \bgf{\textbf{68.4}} & \bgf{34.0} \\
    & BMF-ICL & \bgf{\textbf{61.7}} & \bgf{\textbf{42.9}}$^{\dagger}$ & \bgf{\textbf{63.0}}$^{\dagger}$ & \bgf{\textbf{67.8}}$^{\dagger}$ & \bgf{\textbf{47.2}} & \bgf{63.7} & \bgf{67.9} & \bgf{\textbf{35.0}}$^{\dagger}$ \\ 
    \hdashline
    \multirow{5}{*}{w/o TL} & Non-ICL & 44.9 & 25.0 & 44.3 & 32.7 & 25.1 & 39.1 & 37.7 & 22.9 \\
    & \citet{winata-etal-2021-language} & - & \bgf{29.5} & \bgf{48.9} & \bgf{\textbf{55.1}} & \bgf{35.5} & \bgf{47.0} & \bgf{40.9} & \bgf{26.8} \\
    & \citet{winata-etal-2022-cross} & \bgf{47.5} & \bgf{30.0} & \bgf{49.0} & \bgf{40.2} & \bgf{35.5} & \bgf{45.7} & \bgf{39.7} & \bgf{25.6} \\
    & \citet{nie-etal-2023-cross} & \bgf{49.0} & \bgf{\textbf{31.3}} & \bgf{48.6} & \bgf{50.3} & \bgf{36.9} & \bgf{49.1} & \bgf{40.3} & \bgf{27.7} \\
    & BMF-ICL & \bgf{\textbf{52.1}}$^{\dagger}$ & \bgf{\textbf{31.3}} & \bgf{\textbf{50.6}}$^{\dagger}$ & \bgf{54.5} & \bgf{\textbf{37.5}} & \bgf{\textbf{53.3}}$^{\dagger}$ & \bgf{\textbf{42.0}}$^{\dagger}$ & \bgf{\textbf{30.3}}$^{\dagger}$ \\
    \bottomrule
  \end{tabular}
  }
  \caption{Aya.}
  \end{subtable}
  \hfill
  \begin{subtable}{\textwidth}
  \centering
  \resizebox{0.9\textwidth}{!}{
  \begin{tabular}{llccccccccc}
    \toprule
    & & en & zh & fr & de & ja & nl & pt & ru \\
    \midrule
    \multirow{4}{*}{w/ TL} & Random-ICL & 79.7 & 65.1 & 79.1 & 83.8 & 78.5 & 83.3 & 79.5 & 55.7 \\
    & \citet{etxaniz-etal-2024-multilingual} & - & \bgf{65.8} & \bgf{80.1} & \bgf{85.1} & \bgfr{77.3} & \bgf{\textbf{85.8}} & \bgf{83.0} & \bgfr{54.5} \\
    & \citet{nguyen-etal-2024-democratizing} & - & \bgf{65.2} & \bgf{79.5} & \bgf{84.5} & \bgf{79.0} & \bgfr{83.1} & \bgf{84.1} & \bgf{56.6}  \\
    & BMF-ICL & \bgf{\textbf{80.4}} & \bgf{\textbf{68.5}}$^{\dagger}$ & \bgf{\textbf{82.3}}$^{\dagger}$ & \bgf{\textbf{86.1}}$^{\dagger}$ & \bgf{\textbf{80.1}}$^{\dagger}$ & \bgf{85.7} & \bgf{\textbf{85.5}}$^{\dagger}$ & \bgf{\textbf{58.0}}$^{\dagger}$ \\
    \hdashline
    \multirow{5}{*}{w/o TL} & Non-ICL & 77.2 & 64.0 & 78.3 & 81.2 & 76.3 & 81.6 & 77.7 & 52.9  \\
    & \citet{winata-etal-2021-language} & - & \bgf{64.1} & \bgf{78.4} & \bgf{82.1} & \bgfr{76.0} &  \bgf{\textbf{83.8}} & \bgfr{77.6} & \bgfr{52.8} \\
    & \citet{winata-etal-2022-cross} & \bgfr{77.0} & \bgfr{63.9} & \bgf{78.5} & \bgfr{81.0} & \bgf{76.5} & \bgf{82.0} & \bgf{78.0} & \bgf{53.3} \\
    & \citet{nie-etal-2023-cross} & \bgf{78.8} & \bgf{64.1} & \bgf{\textbf{78.7}} & \bgf{82.0} & \bgf{77.3} & \bgfr{81.0} & \bgf{78.5} & \bgf{53.3} \\
    & BMF-ICL & \bgf{\textbf{79.2}} & \bgf{\textbf{65.1}}$^{\dagger}$ & \bgf{78.5} & \bgf{\textbf{83.1}}$^{\dagger}$ & \bgf{\textbf{78.4}}$^{\dagger}$ & \bgf{83.5} & \bgf{\textbf{78.8}} & \bgf{\textbf{55.0}}$^{\dagger}$ \\
    \bottomrule
  \end{tabular}
  }
  \caption{GPT-3.5.}
  \end{subtable}
  \hfill
  \begin{subtable}{\textwidth}
  \centering
  \resizebox{0.9\textwidth}{!}{
  \begin{tabular}{llccccccccc}
    \toprule
    & & en & zh & fr & de & ja & nl & pt & ru \\
    \midrule
    \multirow{4}{*}{w/ TL} & Random-ICL & 82.2 & 68.8 & 79.3 & 83.5 & 78.8 & 83.5 & 80.0 & 53.3 \\
    & \citet{etxaniz-etal-2024-multilingual} & - & \bgfr{67.5} & \bgf{79.9} & \bgf{84.7} & \bgfr{76.2} & \bgf{84.8} & \bgf{81.9} & \bgfr{52.9} \\
    & \citet{nguyen-etal-2024-democratizing} & - & \bgf{70.1} & \bgf{81.1} & \bgf{84.3} & \bgfr{78.1} & \bgf{84.0} & \bgf{83.3} & \bgf{53.7} \\
    & BMF-ICL & \bgf{\textbf{83.3}}$^{\dagger}$ & \bgf{\textbf{71.7}}$^{\dagger}$ & \bgf{\textbf{82.5}}$^{\dagger}$ & \bgf{\textbf{85.0}} & \bgf{\textbf{80.8}}$^{\dagger}$ & \bgf{\textbf{86.5}}$^{\dagger}$ & \bgf{\textbf{86.6}}$^{\dagger}$ & \bgf{\textbf{55.8}}$^{\dagger}$ \\
    \hdashline
    \multirow{5}{*}{w/o TL} & Non-ICL & 79.7 & 66.0 & 75.4 & 81.9 & 76.0 & 82.3 & 78.1 & 50.3 \\
    & \citet{winata-etal-2021-language} & - & \bgfr{65.2} & \bgf{76.0} & \bgf{83.0} & \bgfr{75.1} & \bgf{82.8} & \bgf{79.0} & \bgfr{49.7} \\
    & \citet{winata-etal-2022-cross} & \bgfr{78.2} & \bgfr{65.6} & \bgf{75.8} & \bgf{82.3} & \bgf{76.5} & \bgf{82.6} & \bgf{79.0} & \bgf{50.5} \\
    & \citet{nie-etal-2023-cross} & \bgf{\textbf{81.9}} & \bgf{67.7} & \bgf{78.1} & \bgf{\textbf{83.2}} & \bgf{77.0} & \bgf{82.5} &  \bgf{79.2} & \bgf{51.9}  \\
    & BMF-ICL & \bgf{81.3} & \bgf{\textbf{69.0}}$^{\dagger}$ & \bgf{\textbf{79.6}}$^{\dagger}$ & \bgf{\textbf{83.2}} & \bgf{\textbf{78.3}}$^{\dagger}$ & \bgf{\textbf{83.0}} & \bgf{\textbf{80.3}}$^{\dagger}$ & \bgf{\textbf{53.0}}$^{\dagger}$  \\
    \bottomrule
  \end{tabular}
  }
  \caption{GPT-4.}
  \end{subtable}
  \caption{Results for baseline ICL methods and our method on mCSQA across the four LLMs. Red and blue indicate scores lower and higher than Random-ICL or Non-ICL. The top half of each table shows results with the target language in the example pool (``w/ TL''), and the bottom half without (``w/o TL''). $\dagger$ indicates statistically significant differences between the highest and second highest score in each LLM according to McNemar’s test ($p < 0.01$).}
  \label{tbl:mcsqa_result}
\end{table*}

\begin{table*}[t!]
  \begin{subtable}{\textwidth}
  \centering
  \small
  \resizebox{1\textwidth}{!}{
  \begin{tabular}{@{}l@{ ~ }l@{ ~ }c@{ ~ }c@{ ~ }c@{ ~ }c@{ ~ }c@{ ~ }c@{ ~ }c@{ ~ }c@{ ~ }c@{ ~ }c@{ ~ }c@{}}
    \toprule
    & & en & ar & bn & fi & id & ja & sw & ko & ru & te & th \\
    \midrule
    \multirow{4}{*}{w/ TL} & Random-ICL & 60.4 & 59.3 & 55.8 & 68.5 & 65.5 & 64.0 & \textbf{59.4} & 63.7 & 60.1 & 59.4 & 58.4 \\
     & \citet{etxaniz-etal-2024-multilingual} & - & \bgfr{55.3} & \bgfr{53.2} & \bgf{70.5} & \bgf{66.3} & \bgfr{63.6} & \bgfr{58.3} & \bgfr{62.2} & \bgfr{58.8} & \bgfr{56.3} & \bgf{59.0} \\
    & \citet{nguyen-etal-2024-democratizing} & - & \bgfr{58.0} & \bgf{57.0} & \bgf{70.8} & \bgf{67.1} & \bgf{65.0} & \bgfr{57.6} & \bgf{64.0} & \bgfr{59.0} & \bgfr{58.5} & \bgf{59.2} \\
    & BMF-ICL & \bgf{\textbf{62.8}}$^{\dagger}$ & \bgf{\textbf{60.2}} & \bgf{\textbf{58.8}}$^{\dagger}$ & \bgf{\textbf{71.9}}$^{\dagger}$ & \bgf{\textbf{68.3}}$^{\dagger}$ & \bgf{\textbf{65.5}} & \bgfr{59.0} & \bgf{\textbf{65.2}}$^{\dagger}$ & \bgf{\textbf{60.7}} & \bgf{\textbf{60.1}} & \bgf{\textbf{60.4}}$^{\dagger}$ \\
    \hdashline
    \multirow{5}{*}{w/o TL} & Non-ICL & 39.3 & 43.3 & 39.1 & 40.9 & 39.3 & 42.6 & 37.0 & 41.4 & 35.5 & 40.8 & 37.5 \\
    & \citet{winata-etal-2021-language} & - & \bgf{44.9} & \bgf{41.6} & \bgf{\textbf{48.8}} & \bgf{43.6} & \bgf{44.9} & \bgf{43.6} & \bgf{45.9} & \bgf{37.3} & \bgf{43.9} & \bgf{39.2} \\
    & \citet{winata-etal-2022-cross} & \bgf{45.9} & \bgf{47.7} & \bgf{43.7} & \bgf{47.5} & \bgf{45.7} & \bgf{45.3} & \bgf{44.8} & \bgf{45.2} & \bgf{38.9} & \bgf{44.7} & \bgf{39.6} \\
    & \citet{nie-etal-2023-cross} & \bgf{47.0} & \bgf{49.0} & \bgf{46.8} & \bgf{48.0} & \bgf{47.1} & \bgf{48.9} & \bgf{47.0} & \bgf{47.3} & \bgf{39.0} & \bgf{45.7} & \bgf{40.1} \\
    & BMF-ICL & \bgf{\textbf{49.8}}$^{\dagger}$ & \bgf{\textbf{51.1}}$^{\dagger}$ & \bgf{\textbf{50.7}}$^{\dagger}$ & \bgf{48.5} & \bgf{\textbf{50.3}}$^{\dagger}$ & \bgf{\textbf{50.9}}$^{\dagger}$ & \bgf{\textbf{48.6}}$^{\dagger}$ & \bgf{\textbf{49.9}}$^{\dagger}$ & \bgf{\textbf{45.7}}$^{\dagger}$ & \bgf{\textbf{49.2}}$^{\dagger}$ & \bgf{\textbf{41.0}} \\
    \bottomrule
  \end{tabular}
  }
  \caption{BLOOMZ.}
  \end{subtable}
  \hfill
  \begin{subtable}{\textwidth}
  \centering
  \resizebox{1\textwidth}{!}{
  \begin{tabular}{@{}l@{ ~ }l@{ ~ }c@{ ~ }c@{ ~ }c@{ ~ }c@{ ~ }c@{ ~ }c@{ ~ }c@{ ~ }c@{ ~ }c@{ ~ }c@{ ~ }c@{}}
    \toprule
    & & en & ar & bn & fi & id & ja & sw & ko & ru & te & th \\
    \midrule
    \multirow{4}{*}{w/ TL} & Random-ICL & 59.3 & 57.0 & 56.3 & 67.0 & 62.0 & 63.4 & 58.7 & \textbf{61.7} & 59.3 & 60.1 & 54.3 \\
    & \citet{etxaniz-etal-2024-multilingual} & - & \bgf{56.3} & \bgfr{55.5} & \bgf{69.1} & \bgfr{61.1} & \bgf{62.2} & \bgfr{56.0} & \bgfr{60.3} & \bgf{60.1} & \bgfr{53.1} & \bgfr{52.7} \\
    & \citet{nguyen-etal-2024-democratizing} & - & \bgf{57.8} & \bgfr{56.0} & \bgf{70.1} & \bgf{62.4} & \bgfr{63.1} & \bgfr{57.3} & \bgfr{61.1} & \bgf{59.6} & \bgfr{53.9} & \bgfr{54.0} \\
    & BMF-ICL & \bgf{\textbf{62.5}}$^{\dagger}$ & \bgf{\textbf{59.9}}$^{\dagger}$ & \bgf{\textbf{59.5}}$^{\dagger}$ & \bgf{\textbf{71.5}}$^{\dagger}$ & \bgf{\textbf{63.5}}$^{\dagger}$ & \bgf{\textbf{65.8}}$^{\dagger}$ & \bgf{\textbf{60.0}}$^{\dagger}$ & \bgfr{61.2} & \bgf{\textbf{60.9}} & \bgf{\textbf{61.3}}$^{\dagger}$ & \bgf{\textbf{58.9}}$^{\dagger}$ \\
    \hdashline
    \multirow{5}{*}{w/o TL} & Non-ICL & 40.5 & 47.5 & 31.5 & 45.6 & 37.8 & 33.9 & 36.6 & 45.7 & 38.0 & 37.5 & 40.1 \\
    & \citet{winata-etal-2021-language} & - & \bgf{50.2} & \bgf{32.5} & \bgf{48.4} & \bgf{40.3} & \bgf{35.4} & \bgf{37.8} & \bgf{46.0} & \bgf{39.7} & \bgf{39.0} & \bgf{41.4} \\
    & \citet{winata-etal-2022-cross} & \bgf{45.1} & \bgf{53.3} & \bgf{33.1} & \bgf{47.0} & \bgf{43.9} & \bgf{36.1} & \bgf{39.1} & \bgf{46.7} & \bgf{40.5} & \bgf{40.4} & \bgf{42.5} \\
    & \citet{nie-etal-2023-cross} & \bgf{49.0} & \bgf{51.7} & \bgf{34.2} & \bgf{48.2} & \bgf{44.7} & \bgf{36.6} & \bgf{39.6} &\bgf{48.8} & \bgf{50.1} & \bgf{40.0} & \bgf{42.2} \\
    & BMF-ICL & \bgf{\textbf{50.6}}$^{\dagger}$ & \bgf{\textbf{55.4}}$^{\dagger}$ &  \bgf{\textbf{40.2}}$^{\dagger}$ & \bgf{\textbf{52.6}}$^{\dagger}$ & \bgf{\textbf{46.8}}$^{\dagger}$ & \bgf{\textbf{40.1}}$^{\dagger}$ & \bgf{\textbf{40.3}} & \bgf{\textbf{50.4}}$^{\dagger}$ & \bgf{\textbf{53.8}}$^{\dagger}$ & \bgf{\textbf{46.4}}$^{\dagger}$ & \bgf{\textbf{48.5}}$^{\dagger}$ \\
    \bottomrule
  \end{tabular}
  }
  \caption{Aya.}
  \end{subtable}
  \hfill
  \begin{subtable}{\textwidth}
  \centering
  
  \resizebox{1\textwidth}{!}{
  \begin{tabular}{@{}l@{ ~ }l@{ ~ }c@{ ~ }c@{ ~ }c@{ ~ }c@{ ~ }c@{ ~ }c@{ ~ }c@{ ~ }c@{ ~ }c@{ ~ }c@{ ~ }c@{}}
    \toprule
    & & en & ar & bn & fi & id & ja & sw & ko & ru & te & th \\
    \midrule
    \multirow{5}{*}{w/ TL} & Random-ICL & 75.9 & \textbf{68.2} & 66.8 & 72.3 & 73.8 & 75.6 & 65.4 & 71.9 & 65.3 & 65.8 & 63.3 \\
    & \citet{etxaniz-etal-2024-multilingual} & - & \bgfr{66.3} & \bgfr{65.7} & \bgf{73.2} & \bgf{75.0} & \bgfr{75.0} & \bgfr{65.2} & \bgfr{71.5} & \bgfr{64.2} & \bgfr{64.8} & \bgf{63.5} \\
    & \citet{nguyen-etal-2024-democratizing} & - & \bgfr{67.7} & \bgf{68.0} & \bgf{74.2} & \bgf{74.6} & \bgfr{75.5} & \bgf{65.7} & \bgfr{70.7} & \bgf{65.8} & \bgfr{65.1} & \bgf{64.0} \\
    & BMF-ICL & \bgf{76.1} & \bgfr{68.0} & \bgf{\textbf{69.1}}$^{\dagger}$ & \bgf{\textbf{75.5}}$^{\dagger}$ & \bgf{\textbf{76.0}}$^{\dagger}$ & \bgf{\textbf{77.1}}$^{\dagger}$ & \bgf{\textbf{66.0}} & \bgf{\textbf{72.6}} & \bgf{\textbf{66.9}}$^{\dagger}$ & \bgf{\textbf{66.5}} & \bgf{\textbf{65.8}}$^{\dagger}$ \\
    \hdashline
    \multirow{5}{*}{w/o TL} & Non-ICL & 72.3 & 62.3 & 60.1 & 69.0 & 71.3 & 72.9 & 61.7 & 70.8 & 59.9 & 61.6 & 62.3 \\
    & \citet{winata-etal-2021-language} & - & \bgf{63.3} & \bgf{60.8} & \bgf{70.1} & \bgf{72.8} & \bgf{74.0} & \bgf{62.0} & \bgf{71.3} & \bgf{60.6} & \bgf{62.0} & \bgf{62.9} \\
    & \citet{winata-etal-2022-cross} & \bgfr{70.1} & \bgf{64.0} & \bgf{61.3} & \bgf{69.3} & \bgf{74.1} & \bgf{74.8} & \bgf{62.2} & \bgf{71.5} & \bgf{62.2} & \bgf{62.2} & \bgf{62.6} \\
    & \citet{nie-etal-2023-cross} &  \bgf{72.6} & \bgf{64.2} & \bgf{61.7} & \bgf{70.5} & \bgf{74.3} & \bgf{75.4} & \bgf{\textbf{62.6}} & \bgf{71.8} & \bgf{63.0} & \bgf{62.8} & \bgf{63.0} \\
    & BMF-ICL & \bgf{\textbf{73.0}} &  \bgf{\textbf{68.9}}$^{\dagger}$ & \bgf{\textbf{62.5}}  & \bgf{\textbf{71.6}}$^{\dagger}$ & \bgf{\textbf{76.9}}$^{\dagger}$ & \bgf{\textbf{76.6}}$^{\dagger}$ & \bgf{\textbf{62.6}} & \bgf{\textbf{72.0}} & \bgf{\textbf{64.6}}$^{\dagger}$ & \bgf{\textbf{63.0}} & \bgf{\textbf{63.1}} \\
    \bottomrule
  \end{tabular}
  }
  \caption{GPT-3.5.}
  \end{subtable}
  \hfill
  \begin{subtable}{\textwidth}
  \centering
  
  \resizebox{1\textwidth}{!}{
  \begin{tabular}{@{}l@{ ~ }l@{ ~ }c@{ ~ }c@{ ~ }c@{ ~ }c@{ ~ }c@{ ~ }c@{ ~ }c@{ ~ }c@{ ~ }c@{ ~ }c@{ ~ }c@{}}
    \toprule
    & & en & ar & bn & fi & id & ja & sw & ko & ru & te & th \\
    \midrule
    \multirow{4}{*}{w/ TL} & Random-ICL & 80.1 & 69.2 & 65.1 & 73.0 & 75.1 & 76.0 & 64.0 & 72.0 & 66.4 & 66.0 & 65.0 \\
    & \citet{etxaniz-etal-2024-multilingual} & - & \bgfr{68.3} & \bgf{65.5}  & \bgf{73.8} & \bgfr{74.7} & \bgfr{75.5} & \bgfr{63.0} & \bgf{72.8} & \bgfr{65.0} & \bgf{66.6} & \bgf{65.5} \\
    & \citet{nguyen-etal-2024-democratizing} & - & \bgf{69.3} & \bgf{\textbf{66.0}} & \bgf{74.0} & \bgf{76.0} & \bgfr{75.9} & \bgfr{63.5} & \bgfr{71.7} & \bgf{\textbf{67.8}} & \bgfr{65.4} & \bgf{66.3} \\
    & BMF-ICL & \bgf{\textbf{80.7}} & \bgf{\textbf{71.5}}$^{\dagger}$ & \bgf{\textbf{66.0}} & \bgf{\textbf{75.7}}$^{\dagger}$ & \bgf{\textbf{76.6}} & \bgf{\textbf{76.4}} & \bgf{\textbf{64.2}} & \bgf{\textbf{74.8}}$^{\dagger}$ & \bgf{67.6} & \bgf{\textbf{66.3}} & \bgf{\textbf{67.1}} \\
    \hdashline
    \multirow{5}{*}{w/o TL} & Non-ICL & 77.3 & 63.0 & 62.2 & 70.7 & 74.3 & 72.9 & 62.0 & 70.1 & 62.7 & 64.1 & 61.7 \\
    & \citet{winata-etal-2021-language} & - & \bgf{63.9} & \bgf{64.0} & \bgf{71.6} & \bgf{74.6} & \bgf{73.2}  & \bgf{62.5} & \bgf{71.1} & \bgf{63.9} & \bgf{65.0} & \bgf{62.7} \\
    & \citet{winata-etal-2022-cross} & \bgf{78.6} & \bgf{64.1} & \bgf{64.8} & \bgf{71.2} & \bgf{74.9} & \bgf{74.2} & \bgf{62.8} & \bgf{70.7} & \bgf{64.5} & \bgf{64.7} & \bgf{62.2} \\
    & \citet{nie-etal-2023-cross} & \bgf{79.1} & \bgf{64.5} & \bgf{65.0} & \bgf{71.1} & \bgf{74.7} & \bgf{74.1} & \bgf{63.1} & \bgf{71.3} & \bgf{64.0} & \bgf{65.3} & \bgf{63.1} \\
    & BMF-ICL & \bgf{\textbf{80.5}}$^{\dagger}$ & \bgf{\textbf{66.8}}$^{\dagger}$ & \bgf{\textbf{66.1}}$^{\dagger}$ & \bgf{\textbf{72.5}} & \bgf{75.1} & \bgf{\textbf{75.9}}$^{\dagger}$ & \bgf{\textbf{63.7}} & \bgf{\textbf{71.8}} & \bgf{\textbf{65.7}}$^{\dagger}$ & \bgf{\textbf{65.8}} & \bgf{\textbf{64.5}}$^{\dagger}$ \\
    \bottomrule
  \end{tabular}
  }
  \caption{GPT-4.}
  \end{subtable}
  \caption{Results for baseline ICL methods and our method on TYDI across the four LLMs. Red and blue indicate scores lower and higher than Random-ICL or Non-ICL. The top half of each table shows results with the target language in the example pool (``w/ TL''), and the bottom half without (``w/o TL''). $\dagger$ indicates statistically significant differences between the highest and second highest score in each LLM according to McNemar’s test ($p < 0.01$).}
  \label{tbl:tydi_result}
\end{table*}

\paragraph{Baseline}

We evaluate our approach against a range of baselines under two ICL settings: one where the example candidates include the target language, and one where they do not.
In both settings, we use 8 examples for ICL, consistent with our proposed method.

\begin{itemize}[leftmargin=*, itemsep=0pt, topsep=0pt, parsep=0pt, partopsep=0pt]
\item \textbf{With target-language examples:} 
    \begin{itemize} 
        \item \textit{Random-ICL}: We randomly select 8 examples (source text and reference text) from the target language candidate set. 
        \item \citet{etxaniz-etal-2024-multilingual}: We translate both the input and examples into English, which is dominant in the model’s training data,  and feeds them into the MLLM for ICL.\footnote{Following the original translation setting, we use four examples from the FLORES-200 dataset~\cite{costa2022no}, prepending each sentence with its language name (e.g., \textit{English: Mary did not slap the green witch.}).} 
        \item \citet{nguyen-etal-2024-democratizing}: This baseline generates pseudo-reference texts in the target language by leveraging examples from a high-resource language, then pairs them with the original target-language source text to create ICL examples.\footnote{Following their approach, we use English as the high-resource language and randomly sample examples from the candidate set. }
    \end{itemize}
\item \textbf{Without target-language examples:} 
    \begin{itemize} 
        \item \textit{Non-ICL}: A zero-shot baseline that provides only the input text (no examples). 
        \item \citet{winata-etal-2021-language}: This baseline provides English examples to the MLLM while performing inference on the input in the target language. We randomly sample examples from the English candidate set. 
        \item \citet{winata-etal-2022-cross}: We randomly select examples for ICL from a pool across various languages excluding the target language.
        \item \citet{nie-etal-2023-cross}: We use XLM-based~\citep{conneau-etal-2020-unsupervised} multilingual sentence embeddings to select examples in high-resource languages similar to the input text in a low-resource language. Following previous work, we use English, German, and Chinese as high-resource languages for mCSQA, and English and Arabic for TYDI. 
    \end{itemize} 
\end{itemize}

\subsection{Experimental Results}

\autoref{tbl:mcsqa_result} and \autoref{tbl:tydi_result} present the test set results on the mCSQA and TYDI datasets, for the four LLMs.
The upper half of each table represents the setting where instances of the target language are included in the example candidates, while the lower half represents the setting where they are not included.

Comparisons with baselines show that BMF-ICL consistently achieves the best performance on mCSQA (in \textbf{28/32} cases) when target-language examples are included and also (in \textbf{27/32} cases) when they are not. 
On TYDI, it attains the highest performance in \textbf{39/44} cases under the target-language-included setting and in \textbf{42/44} cases under the target-language-excluded setting. 
These findings confirm that BMF-ICL achieves state-of-the-art performance overall. 
Its improvements are observed across all four tested LLMs, indicating that BMF-ICL is not heavily model-dependent.
Furthermore, BMF-ICL does not exhibit poor performance on any particular language. 
Notably, the method is more effective in cases where the target language is not included among the example candidates.
In contrast, existing approaches are not always better than the Random-ICL or Non-ICL baselines, particularly in the target-language-included setting, underscoring the challenge of achieving stable improvements without quantitative optimization.

\section{Analysis}
\label{sec:ana}

\subsection{Ablation Study}

\begin{figure}[t!]
    \centering
    \begin{subfigure}{0.47\textwidth} 
        \centering
        \includegraphics[width=\textwidth]{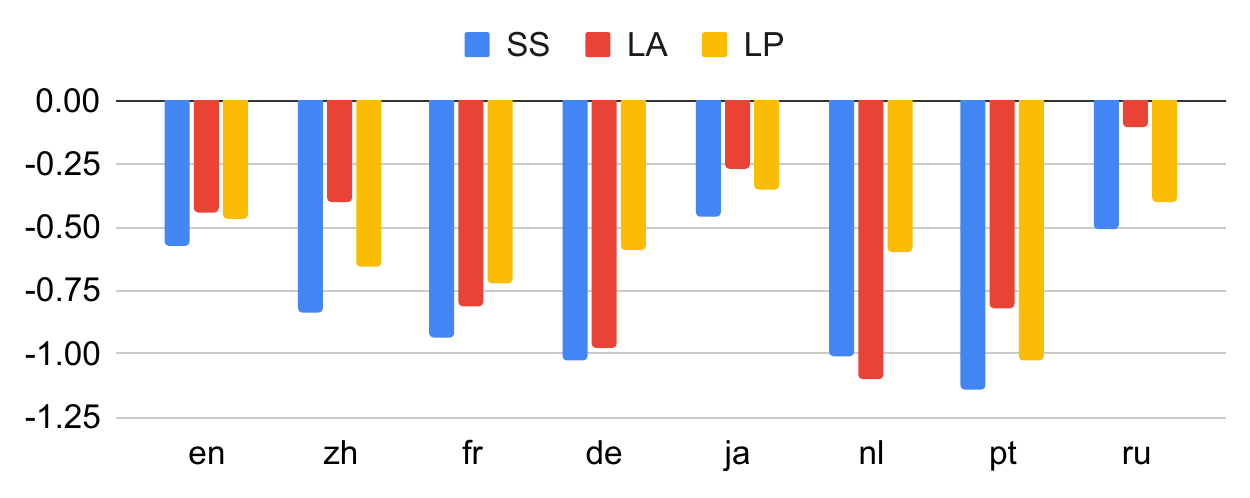} 
        \caption{Accuracy variations on the mCSQA dataset.}
        \label{fig:mcsqa_ab}
    \end{subfigure}

    \vspace{1em}
    \begin{subfigure}{0.47\textwidth} 
        \centering
        \includegraphics[width=\textwidth]{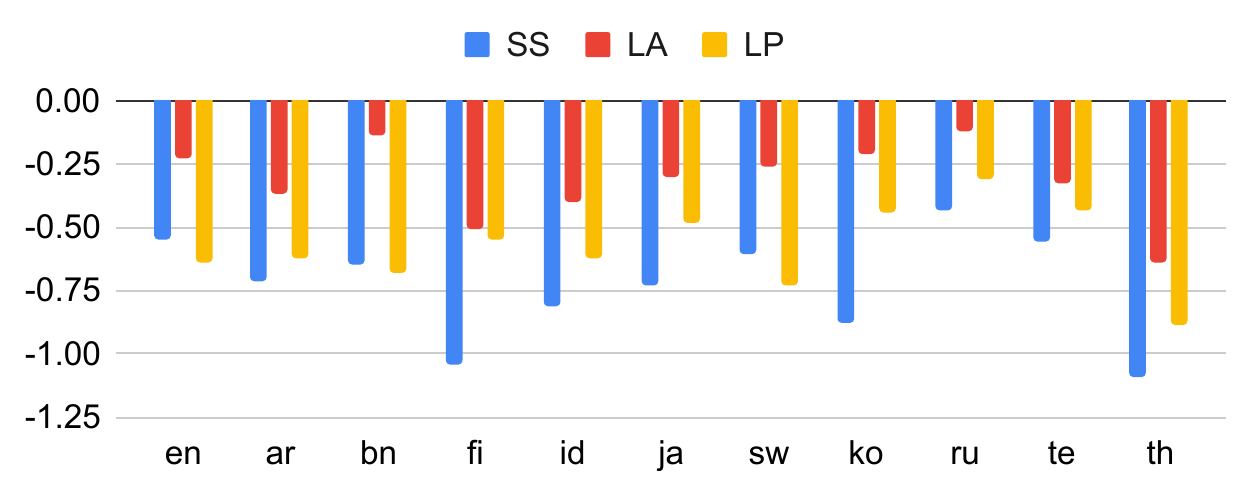}
        \caption{Accuracy variations on the TYDI dataset.}
        \label{fig:tydi_ab}
    \end{subfigure}
    \caption{Results of the ablation study. Semantic similarity, linguistic alignment, and language-specific performance are denoted as SS, LA, and LP, respectively.}
    \label{fig:ablation}
\end{figure}
To assess the importance of each factor in BNF-ICL, we conducted an ablation study by removing one factor at a time. 
For example, when evaluating the impact of semantic similarity, we set $\alpha=0$ in \autoref{eq:total} while keeping $\beta$ and $\gamma$ unchanged.\footnote{The results with all weights set to 1 are shown in \autoref{apx:sec:uni_weights}.}
\autoref{fig:ablation} displays the ablation results under the setting where the target language is included among the candidate examples for both mCSQA and TYDI, averaged over the four MLLMs.

Across all languages, performance declines whenever one of the three factors is excluded, indicating the importance of jointly considering all three.
In mCSQA, 7 out of 8 languages, and in TYDI, 8 out of 11 languages, exhibited the  largest drop when the semantic similarity factor was removed.
This suggests that, as in monolingual settings~\cite{liu-etal-2022-makes}, providing semantically similar examples contributes to performance improvement in multilingual settings.
Additionally, within closely-related language groups such as en--de--nl or fr--pt, linguistic alignment plays a crucial role leading to pronounced performance declines compared to other languages when removed.

\subsection{The Diversity of Languages in the Examples}

\begin{table}[t]
\centering
\scriptsize
\begin{tabular}{lcccccccccc}
\toprule
& \multicolumn{4}{c}{mCSQA} & \multicolumn{4}{c}{TYDI} \\
\cmidrule(lr){2-5} \cmidrule(lr){6-9}
& All & SS & LA & LP 
& All & SS & LA & LP \\
\midrule
1 & 0.05 & 0.00 & 0.10 & 0.15 & 0.01 & 0.01 & 0.01 & 0.05 \\
2 & 0.19 & 0.01 & \textbf{0.34} & \textbf{0.31} & 0.14 & 0.13 & 0.11 & 0.17 \\
3 & \textbf{0.31} & \textbf{0.40} & 0.23 & 0.27 & 0.28 & 0.18 & \textbf{0.30} & \textbf{0.29} \\
4 & 0.24 & 0.32 & 0.16 & 0.19 & \textbf{0.30} & 0.20 & 0.28 & 0.23 \\
5 & 0.11 & 0.17 & 0.07 & 0.05 & 0.13 & \textbf{0.29} & 0.20 & 0.10 \\
6 & 0.06 & 0.07 & 0.06 & 0.02 & 0.08 & 0.10 & 0.08 & 0.09 \\
7 & 0.03 & 0.03 & 0.03 & 0.01 & 0.05 & 0.07 & 0.01 & 0.06 \\
8 & 0.01 & 0.00 & 0.01 & 0.00 & 0.01 & 0.02 & 0.01 & 0.01 \\
\bottomrule
\end{tabular}
\caption{The proportion of instances with each language type count within the 8 examples of BMF-ICL. \textit{All} represents the results of a method that considers three factors with optimized weights. SS, LA, and LP represent results considering only semantic similarity, linguistic alignment, and language-specific performance, respectively.}
\label{tbl:lang_type}
\end{table}

BMF-ICL demonstrates improved performance through cross-linguistic knowledge transfer, as clarified in our study.
For further analysis, we examined how many distinct language types are included among the 8 examples in each ICL prompt. 
Based on these counts, we then computed the proportion of test instances associated with each distinct language-type count.
We report results for two settings: (i) BMF-ICL optimized with all three factors, and (ii) ablation settings where only one factor is set to 1 in Eq.~\eqref{eq:total}, while the others are set to 0, allowing us to assess each factor's impact on language diversity.

Table~\ref{tbl:lang_type} shows the distribution of distinct language-type counts for examples selected by BMF-ICL when the target language is included among the candidates. 
The \emph{All} column represents results considering all weights, while SS, LA, and PL correspond to each factor used in isolation.
Under \emph{All}, the most frequently selected diversity level is three languages in mCSQA and four in TYDI, indicating that BMF-ICL naturally favors multilingual examples. 
Furthermore, the semantic similarity factor (SS) tends to yield higher language diversity than the other two factors, underscoring its particular importance for encouraging more varied language selection.

\section{Related Work}
\label{sec:rel}

\citet{etxaniz-etal-2024-multilingual} showed that translating low-resource language inputs and examples into English improves LLM performance compared to direct inference in the original language. This approach leverages the English-centric training of most LLMs, but may not fully capture linguistic, cultural, or societal norms. Additionally, using translated examples for ICL risks information loss or distortion, as LLMs struggle with accurately conveying cultural or societal nuances~\cite{yao2023benchmarking,tenzer2024ai,intrator-etal-2024-breaking}.

\citet{winata-etal-2021-language} discovered that providing English examples for ICL improves LLM inference for both English and non-English tasks, though English was heuristically chosen. \citet{winata-etal-2022-cross} showed that randomly sampling from a multilingual dataset outperforms selecting examples based on geographical or linguistic proximity. However, the role of semantic alignment and language-specific capacity in example selection remains unclear in the original work.

\citet{nie-etal-2023-cross} introduced a method that uses multilingual sentence embeddings~\cite{conneau-etal-2020-unsupervised} to select examples in high-resource language similar to the input text in low-resource language.
The multilingual sentence embeddings do not explicitly distinguish between semantic and linguistic similarity, making it impossible to adjust their optimal balance for ICL examples.
Moreover, this study focuses on only masked language models such as mBERT~\cite{devlin-etal-2019-bert} and XLM~\cite{conneau-etal-2020-unsupervised} rather than LLMs.

To leverage unlabeled datasets in low-resource languages, \citet{nguyen-etal-2024-democratizing} address the data scarcity in low-resource languages  using instances from diverse high-resource languages as ICL examples to create synthetic data from unlabeled datasets in low-resource languages, which are then used as ICL examples in a low-resource setting.
However, this method does not consider the similarity between the input and example texts.

The following studies have proposed MICL methods specialized for binary classification tasks.
\citet{tanwar-etal-2023-multilingual} proposed a method that uses multilingual sentence embeddings~\cite{reimers-gurevych-2020-making} to retrieve similar texts in another language as examples for ICL in a cross-lingual setting.
This method explicitly presents cross-lingual label correspondences (e.g., \textit{In French, ``bad'' means ``mal''}).
\citet{cahyawijaya-etal-2024-llms} introduced query alignment for ICL, selecting examples from parallel data with source texts that match the input language and target texts in high-resource languages.
This method used multilingual sentence embeddings~\cite{reimers-gurevych-2019-sentence,reimers-gurevych-2020-making} to measure the similarity between the input text and the source texts in the parallel data, selecting semantically similar texts as examples.
The labels from the high-resource language are used directly, avoiding translation errors.
Unlike these existing studies, which focus on binary classification tasks, our study applies ICL methods to more general generative tasks.

\citet{qin-etal-2023-cross} introduced a method that processes inputs in languages other than English by using the prompt \textit{Let’s think in English step by step!} to enable step-by-step reasoning in English.
This method consistently improves the performance in languages other than English.
\citet{shi2023language} also demonstrates that step-by-step reasoning enhances the multilingual capabilities of MLLMs.
Unlike our research, which focuses on multilingual knowledge transfer through examples in ICL, this study emphasizes multilingual knowledge transfer within the reasoning process.

\section{Conclusion}

In this paper, we proposed \textbf{balanced multi-factor ICL (BMF-ICL)}, an approach for multilingual example selection in ICL for MLLMs~\cite{Scao2022BLOOMA1,ustun2024aya}.
BMF-ICL quantifies and balances three key factors: semantic similarity, linguistic alignment, and language-specific performance.
By leveraging LaBSE~\cite{feng-etal-2022-language} embeddings for semantic similarity, lang2vec~\cite{littell-etal-2017-uriel} for linguistic alignment, and MLLM likelihoods for language-specific performance, BMF-ICL optimally selects examples through a weighted scoring mechanism. 

Experimental results on the mCSQA and TYDI datasets, using four different MLLMs, demonstrated that BMF-ICL consistently achieves higher accuracy than existing methods, underscoring the critical role of considering all three factors in MICL.
Furthermore, our analysis revealed that in over 95\% of cases, BMF-ICL selects examples spanning multiple languages, highlighting the benefits of cross-lingual knowledge transfer.

\section*{Limitations}
\label{sec:lim}

We demonstrated the effectiveness of the proposed method by conducting large-scale experiments in various languages; however, this does not guarantee performance improvements in all languages.
As future work, it would be worthwhile to validate the method on a broader range of tasks beyond question answering.
On the other hand, since there are not many multilingual datasets created from scratch for each language, this is an aspect that needs to be considered from the dataset creation stage.

\section*{Ethical Considerations}
\label{sec:ethics}

mCSQA~\cite{sakai-etal-2024-mcsqa} is a dataset that reflects common sense across different cultures, and our experimental results indicate that the proposed method enhances the understanding of common sense within each culture by leveraging multilingual information.  
Therefore, it also has the potential to positively impact safety-related tasks such as social biases, morality, and ethics~\cite{kaneko2022gender,kaneko2024eagle,kaneko2024little,anantaprayoon2023evaluating,haemmerl-etal-2023-speaking}, where multicultural factors play a significant role.

\bibliography{custom}

\clearpage
\appendix

\begin{table*}[!t]
\scriptsize
\centering
\begin{tabular}{lccccccc}
\toprule
 & & \multicolumn{3}{c}{\textbf{mCSQA}} & \multicolumn{3}{c}{\textbf{TYDI}} \\
 & Language group & \textbf{Train} & \textbf{Valid} & \textbf{Test} & \textbf{Train} & \textbf{Valid} & \textbf{Test} \\
\midrule
Arabic (ar) & Semitic & - & - & - & 23.0k & 1.3k & 1.4k \\ 
Bengali (bn) & Indo-Aryan & - & - & - & 10.7k & 0.3k & 0.3k \\
Chinese (zh) & Sinitic & 12.2k & 1.5k & 1.5k & - & - & - \\
English (en) & Germanic & 10.9k & 1.3k & 1.3k & 9.2k & 1.0k & 1.0k \\
Finnish (fi) & Finnic & - & - & - & 15.2k & 2.0k & 2.0k \\
French (fr) & Romance & 8.0k & 1.0k & 1.0k & - & - & - \\
German (de) & Germanic & 12.5k & 1.5k & 1.5k & - & - & - \\
Indonesian (id) & Malayo-Polynesian & - & - & - & 14.9k & 1.8k & 1.8k \\ 
Japanese (ja) & Japonic & 11.7k & 1.4k & 1.4k & 16.2k & 1.7k & 1.7k \\
Kiswahili (sw) & Bantu & - & - & - & 17.6k & 2.2k & 2.2k \\
Korean (ko) & Koreanic & - & - & - & 10.9k & 1.6k & 1.7k \\
Dutch (nl) & Germanic & 12.2k & 1.5k & 1.5k & - & - & -  \\
Portuguese (pt) & Romance & 12.7k & 1.5k & 1.5k & - & - & -  \\
Russian (ru) & Slavic & 6.6k & 0.8k & 0.8k & 12.8k & 1.6k & 1.6k \\
Telugu (te) & Dravidian & - & - & - & 24.5k & 2.4k & 2.5k  \\
Thai (th) & Tai & - & - & - & 11.3k & 2.2k & 2.2k \\
\bottomrule
\end{tabular}
\caption{Dataset statistics and language groups for mCSQA and TYDI.}
\label{apx:tbl:data}
\end{table*}

\section{Dataset Statistics}
\label{apx:sec:dataset}

\autoref{apx:tbl:data} shows the data size and language groups for mCSQA~\cite{sakai-etal-2024-mcsqa} and TYDI~\cite{Clark2020TyDiQA}.

\section{Prompts}
\label{apx:sec:prompt}

The following are four candidate prompts for mCSQA.

\begin{tcolorbox}[fontupper=\ttfamily, title={Prompt 1 for mCSQA}]
  \scriptsize
  Answer the question.

  Question: [Question of Example 1]

  a. [Choice A of Example 1]
  
  b. [Choice B of Example 1]
  
  c. [Choice C of Example 1]
  
  d. [Choice D of Example 1]
  
  e. [Choice E of Example 1]

  Answer: [Answer of Example 1]
  
  \vdots

  Question: [Question of Example 8]

  a. [Choice A of Example 8]
  
  b. [Choice B of Example 8]
  
  c. [Choice C of Example 8]
  
  d. [Choice D of Example 8]
  
  e. [Choice E of Example 8]

  Answer: [Answer of Example 8]

  Question: [Question of Input]

  a. [Choice A of Input]
  
  b. [Choice B of Input]
  
  c. [Choice C of Input]
  
  d. [Choice D of Input]
  
  e. [Choice E of Input]

  Answer: 
\end{tcolorbox}

\begin{tcolorbox}[fontupper=\ttfamily, title={Prompt 2 for mCSQA}]
  \scriptsize
  Provide a response to the question.

  Question: [Question of Example 1]

  a. [Choice A of Example 1]
  
  b. [Choice B of Example 1]
  
  c. [Choice C of Example 1]
  
  d. [Choice D of Example 1]
  
  e. [Choice E of Example 1]

  Answer: [Answer of Example 1]
  
  \vdots

  Question: [Question of Example 8]

  a. [Choice A of Example 8]
  
  b. [Choice B of Example 8]
  
  c. [Choice C of Example 8]
  
  d. [Choice D of Example 8]
  
  e. [Choice E of Example 8]

  Answer: [Answer of Example 8]

  Question: [Question of Input]

  a. [Choice A of Input]
  
  b. [Choice B of Input]
  
  c. [Choice C of Input]
  
  d. [Choice D of Input]
  
  e. [Choice E of Input]

  Answer: 
\end{tcolorbox}

\begin{tcolorbox}[fontupper=\ttfamily, title={Prompt 3 for mCSQA}]
  \scriptsize
  Please answer the question.

  Question: [Question of Example 1]

  a. [Choice A of Example 1]
  
  b. [Choice B of Example 1]
  
  c. [Choice C of Example 1]
  
  d. [Choice D of Example 1]
  
  e. [Choice E of Example 1]

  Answer: [Answer of Example 1]
  
  \vdots

  Question: [Question of Example 8]

  a. [Choice A of Example 8]
  
  b. [Choice B of Example 8]
  
  c. [Choice C of Example 8]
  
  d. [Choice D of Example 8]
  
  e. [Choice E of Example 8]

  Answer: [Answer of Example 8]

  Question: [Question of Input]

  a. [Choice A of Input]
  
  b. [Choice B of Input]
  
  c. [Choice C of Input]
  
  d. [Choice D of Input]
  
  e. [Choice E of Input]

  Answer: 
\end{tcolorbox}

\begin{tcolorbox}[fontupper=\ttfamily, title={Prompt 4 for mCSQA}]
  \scriptsize
  Respond to the question.

  Question: [Question of Example 1]

  a. [Choice A of Example 1]
  
  b. [Choice B of Example 1]
  
  c. [Choice C of Example 1]
  
  d. [Choice D of Example 1]
  
  e. [Choice E of Example 1]

  Answer: [Answer of Example 1]
  
  \vdots

  Question: [Question of Example 8]

  a. [Choice A of Example 8]
  
  b. [Choice B of Example 8]
  
  c. [Choice C of Example 8]
  
  d. [Choice D of Example 8]
  
  e. [Choice E of Example 8]

  Answer: [Answer of Example 8]

  Question: [Question of Input]

  a. [Choice A of Input]
  
  b. [Choice B of Input]
  
  c. [Choice C of Input]
  
  d. [Choice D of Input]
  
  e. [Choice E of Input]

  Answer: 
\end{tcolorbox}

The following are four candidate prompts for TYDI.

\begin{tcolorbox}[fontupper=\ttfamily, title={Prompt 1 for TYDI}]
  \scriptsize
  Answer the question using the context.
  
  Context: [Context of Example 1]

  Question: [Question of Example 1]

  Answer: [Answer of Example 1]

  Context: [Context of Example 2]

  Question: [Question of Example 2]

  Answer: [Answer of Example 2]
  
  \vdots

  Context: [Context of Example 7]

  Question: [Question of Example 7]

  Answer: [Answer of Example 7]

  Context: [Context of Example 8]

  Question: [Question of Example 8]

  Answer: [Answer of Example 8]

  Context: [Context of Input]

  Question: [Question of Input]

  Answer: 
\end{tcolorbox}

\begin{tcolorbox}[fontupper=\ttfamily, title={Prompt 2 for TYDI}]
  \scriptsize
  Provide an answer to the question based on the context.
  
  Context: [Context of Example 1]

  Question: [Question of Example 1]

  Answer: [Answer of Example 1]

  Context: [Context of Example 2]

  Question: [Question of Example 2]

  Answer: [Answer of Example 2]
  
  \vdots

  Context: [Context of Example 7]

  Question: [Question of Example 7]

  Answer: [Answer of Example 7]

  Context: [Context of Example 8]

  Question: [Question of Example 8]

  Answer: [Answer of Example 8]

  Context: [Context of Input]

  Question: [Question of Input]

  Answer: 
\end{tcolorbox}

\begin{tcolorbox}[fontupper=\ttfamily, title={Prompt 3 for TYDI}]
  \scriptsize
  Please give an answer to the question using the provided context.
  
  Context: [Context of Example 1]

  Question: [Question of Example 1]

  Answer: [Answer of Example 1]

  Context: [Context of Example 2]

  Question: [Question of Example 2]

  Answer: [Answer of Example 2]
  
  \vdots

  Context: [Context of Example 7]

  Question: [Question of Example 7]

  Answer: [Answer of Example 7]

  Context: [Context of Example 8]

  Question: [Question of Example 8]

  Answer: [Answer of Example 8]

  Context: [Context of Input]

  Question: [Question of Input]

  Answer: 
\end{tcolorbox}

\begin{tcolorbox}[fontupper=\ttfamily, title={Prompt 4 for TYDI}]
  \scriptsize
  Please answer the question by utilizing the context.
  
  Context: [Context of Example 1]

  Question: [Question of Example 1]

  Answer: [Answer of Example 1]

  Context: [Context of Example 2]

  Question: [Question of Example 2]

  Answer: [Answer of Example 2]
  
  \vdots

  Context: [Context of Example 7]

  Question: [Question of Example 7]

  Answer: [Answer of Example 7]

  Context: [Context of Example 8]

  Question: [Question of Example 8]

  Answer: [Answer of Example 8]

  Context: [Context of Input]

  Question: [Question of Input]

  Answer: 
\end{tcolorbox}

\section{The Weights of the Three Factors}
\label{apx:sec:weights}

\begin{table}[t!]
\small
\centering
\begin{subtable}[t]{0.5\textwidth}
\centering
\begin{tabular}{lccccccc}
\toprule
& $\alpha$ & $\beta$ & $\gamma$ \\
\midrule
Chinese & 0.4 & 0.4 & 0.2 \\
English & 0.5 & 0.3 & 0.2 \\
French & 0.4 & 0.3 & 0.3 \\
German & 0.4 & 0.5 & 0.1 \\ 
Japanese & 0.4 & 0.3 & 0.3 \\
Dutch & 0.4 & 0.4 & 0.2 \\
Portuguese & 0.6 & 0.1 & 0.3 \\
Russian & 0.5 & 0.3 & 0.2 \\
\bottomrule
\end{tabular}
\caption{mCSQA.}
\label{apx:tbl:mcsqa_w}
\end{subtable}

\vspace{1cm} 

\begin{subtable}[t]{0.5\textwidth}
\centering
\begin{tabular}{lccccccc}
\toprule
 & $\alpha$ & $\beta$ & $\gamma$ \\
\midrule
Arabic & 0.5 & 0.4 & 0.1 \\ 
Bengali & 0.4 & 0.1 & 0.5 \\
English & 0.6 & 0.3 & 0.1 \\
Finnish & 0.4 & 0.3 & 0.3 \\
Indonesian & 0.4 & 0.2 & 0.4 \\ 
Japanese & 0.6 & 0.2 & 0.2 \\
Kiswahili & 0.4 & 0.4 & 0.2 \\
Korean & 0.5 & 0.4 & 0.1 \\
Russian & 0.5 & 0.4 & 0.1 \\
Telugu & 0.4 & 0.2 & 0.4 \\
Thai & 0.4 & 0.2 & 0.4 \\
\bottomrule
\end{tabular}
\caption{TYDI.}
\label{apx:tbl:tydi_w}
\end{subtable}
\caption{The weights for semantic similarity (SS) with weight $\alpha$, linguistic alignment (LA) with weight $\beta$, and language-specific performance (LP) with weight $\gamma$ in BMF-ICL for each language in the mCSQA and TYDI datasets.}
\label{apx:tbl:weights}
\end{table}

\autoref{apx:tbl:weights} shows weights for semantic similarity with weight $\alpha$, linguistic alignment with weight $\beta$, and language-specific performance with weight $\gamma$ in \autoref{eq:total} in BMF-ICL for each language in the mCSQA and TYDI datasets.

\section{BMF-ICL with Uniform Weights}
\label{apx:sec:uni_weights}

\autoref{apx:tbl:uni_weights} shows the extent to which performance degrades when the weights $\alpha$, $\beta$, and $\gamma$ of BMF-ICL are all set to 1, difference compared to the performance of MBF-ICL with optimized weights.
From the experimental results, it can be observed that optimizing the weights across all settings contributes to performance improvement.

\begin{table}[t!]
\small
\centering
\begin{subtable}[t]{0.5\textwidth}
\centering
\begin{tabular}{lccccccc}
\toprule
& BLOOMZ & Aya & GPT-3.5 & GPT-4 \\
\midrule
Chinese & -5.2 & -3.2 & -2.4 & -2.3 \\
English & -5.4 & -4.2 & -3.0 & -2.6 \\
French & -4.9 & -3.3 & -2.7 & -2.5 \\
German &  -4.6 & -3.3 & -2.3 & -1.9 \\ 
Japanese & -5.0 & -2.9 & -2.6 & -2.6 \\
Dutch & -4.4 & -2.6 & -2.4 & -2.1 \\
Portuguese & -5.2 & -3.0 & -2.2 & -2.1 \\
Russian & -4.7 & -3.6 & -2.3 & -1.8 \\
\bottomrule
\end{tabular}
\caption{mCSQA.}
\label{apx:tbl:uni_mcsqa}
\end{subtable}

\vspace{1cm}

\begin{subtable}[t]{0.5\textwidth}
\centering
\begin{tabular}{lccccccc}
\toprule
 & BLOOMZ & Aya & GPT-3.5 & GPT-4 \\
\midrule
Arabic & -6.6 & -5.1 & -3.9 & -3.2 \\
Bengali & -6.2 & -4.4 & -3.7 & -3.5 \\
English & -.7.0 & -6.6 & -4.8 & -4.3 \\
Finnish & -5.9 & -4.2 & -3.7 & -3.3 \\
Indonesian & -5.5 & -4.7 & -4.0 & -4.1 \\ 
Japanese & -6.3 & -5.3 & -3.6 & -4.0 \\
Kiswahili & -5.2 & -4.7 & -2.9 & -2.7 \\
Korean & -6.1 & -5.2 & -4.0 & -4.3 \\
Russian & -5.9 & -4.6 & -4.1 & -3.6 \\
Telugu & -4.9 & -3.9 & -3.2 & -3.0 \\
Thai & -5.3 & -4.4 & -3.6 & -3.3 \\
\bottomrule
\end{tabular}
\caption{TYDI.}
\label{apx:tbl:uni_tydi}
\end{subtable}
\caption{ in BMF-ICL for each language in the mCSQA and TYDI datasets.}
\label{apx:tbl:uni_weights}
\end{table}

\end{document}